\pdfoutput=1

\documentclass[11pt]{article}

\usepackage[]{acl}

\usepackage{times}
\usepackage{latexsym}

\usepackage[T1]{fontenc}

\usepackage[utf8]{inputenc}

\usepackage{microtype}

\usepackage{ragged2e}
\usepackage{graphicx}
\usepackage{tabularx}
\usepackage{soul}

\usepackage{amsmath}
\usepackage{cuted}
\usepackage{booktabs}
\usepackage{diagbox}
\usepackage{enumitem}

\usepackage{multirow}
\newcolumntype{L}{>{\raggedright\arraybackslash}X}
\newcolumntype{b}{X}
\newcolumntype{s}{>{\hsize=.85\hsize}X}

\title{Multi-Sense Language Modelling}
\begin{document}

\author{Andrea Lekkas$^{1,2,3}$,  Peter Schneider-Kamp$^{1}$, Isabelle Augenstein$^{3}$ \\
$^{1}$Dept. of Mathematics and Computer Science, University of Southern Denmark\\ 
$^{2}$Ordbogen A/S, Odense, Denmark \\
$^{3}$Dept. of Computer Science, University of Copenhagen, Denmark\\
\texttt{gzt740@alumni.ku.dk, petersk@imada.sdu.dk, augenstein@di.ku.dk}}

\maketitle

\begin{abstract}
The effectiveness of a language model is influenced by its token representations, which must encode contextual information and handle the same word form having a plurality of meanings (polysemy).
Currently, none of the common language modelling architectures explicitly model polysemy.
We propose a language model which not only predicts the next word, but also its sense in context. 
We argue that this higher prediction granularity may be useful for end tasks such as assistive writing, and allow for more a precise linking of language models with knowledge bases.
We find that multi-sense language modelling requires architectures that go beyond standard language models, 
and here propose a localized prediction framework that decomposes the task into a word followed by a sense prediction task. To aid sense prediction, we utilise a Graph Attention Network, which encodes definitions and example uses of word senses. 
Overall, we find that multi-sense language modelling is a highly challenging task, and suggest that future work focus on the creation of more annotated training datasets.
\end{abstract}

\section{Introduction}

Any variant of language model, whether standard left-to-right, masked \cite{devlin-etal-2019-bert} or bidirectional \cite{peters-etal-2018-deep} has to address the problem of \textit{polysemy}: the same word form having multiple meanings, as seen in Tab. \ref{tab:example_of_polysemy}. The meaning of a particular occurrence depends on the context, and all modern language modelling architectures from simple RNNs \cite{Mikolov:10} to Transformers \cite{vaswani2017attention} use context-based representations. However, \textit{token} representations in language models are not explicitly disambiguated. 
Single-prototype embeddings, i.e., traditional word vectors, have a 1-to-1 correspondence with word forms. Contextual embeddings change depending on the tokens in their context window, and are employed in recent models like ULMFit \cite{howard-ruder-2018-universal}, ELMo \cite{peters-etal-2018-deep}, and all Transformer architectures. However, even for contextual embeddings, polysemy is handled in an implicit, non-discrete way: the sense that a word assumes in a particular occurrence is unspecified. 

Here, we propose the task of multi-sense language modelling, consisting of not only word, but also sense prediction. We conjecture that multi-sense language modelling would:
\begin{enumerate}[noitemsep,label=\arabic*)]
    \item improve the precision of linking a language model to a knowledge base, as done in \newcite{logan-etal-2019-baracks}, to help generate factually correct language. For instance: ``\textit{The explorers descended in the cave and encountered a bat}'' refers to the entity \textit{`bat (animal)'} and not to \textit{`bat (baseball implement)'}.
    \item be useful in applications such as assistive writing \cite{chen-etal-2012-flow}, where it is desirable to display more information about a word to a user, such as its sense, definition or usage.
\end{enumerate}

\begin{table}[]
\fontsize{10}{10}\selectfont
    \begin{tabular}{p{0.3\columnwidth}p{0.6\columnwidth}} \toprule
        Sentence & Meaning \\ \midrule
        \textit{``John sat on the \textbf{bank} of the river and watched the currents''} & \textbf{bank.n.01}: \textit{sloping land, especially the slope beside a body of water} \\ \vspace*{.1\baselineskip}
        \textit{``Jane went to the \textbf{bank} to discuss the mortgage''} & \vspace*{.1\baselineskip} \textbf{bank.n.02}: \textit{a financial institution that accepts deposits and channels the money into lending activities} \\ \bottomrule
    \end{tabular}
    \caption{Example of polysemy. Senses taken from WordNet 3.0}
    \label{tab:example_of_polysemy}
\end{table}

Another potential use would be to explore if such dictionary information could improve standard language modelling, and reduce the number of training data needed, relevant for e.g. language modelling for low-resource languages.

Consequently, our \textbf{research objectives} are to:
\begin{itemize}[noitemsep]
    \item model next-sense prediction as a task 
    alongside standard next-word prediction in language modelling, and examine the performance of different model architectures. 
    \item encode sense background knowledge from sense definitions and examples, and examine how it can aid sense prediction.
\end{itemize}

As a sense inventory, we use WordNet 3.0 \cite{Miller95wordnet:a}. The sense background knowledge is encoded in a dictionary graph, as shown in Fig. \ref{fig:dictionary_example_graph}. When reading a word \textit{w}, the model can rely on an additional input signal: the state of the node that represents \textit{w} in the dictionary graph (the \textit{``global node''}). Node vectors in the graph are updated by a Graph Attention Network \cite{velickovic2018graph}.

\textbf{Findings}: We find that sense prediction is a significantly more difficult task than standard word prediction. A way to tackle it is to use a localized prediction framework, where the next sense depends on the prediction of the next word.
The most successful model we identified for this uses a hard cut-off for the number of words considered (SelectK, see § \ref{subsec:SelectK}).
The additional input signal from the dictionary graph provides only marginal improvements.
For future work, we argue that multi-sense language modelling would benefit from larger sense-labelled datasets, possibly aided by WordNet super-sense information (categories like food, artifact, person, etc.) to deal with the excessively fine granularity of WordNet senses.

\section{Related Work}

We here briefly discuss relevant works that disambiguate between word senses to address polysemy. They can be grouped in three categories: 1) multi-prototype embeddings not connected to a knowledge base; 2) supervised multi-sense embeddings based on a text corpus that only utilise a KB tangentially as the sense inventory; and 3) models that rely more substantially on features from a KB, like glosses or semantic relations. 

\paragraph{Multi-prototype embeddings} \newcite{huang-etal-2012-improving} learn multi-prototype vectors by clustering word context representations. Single-prototype embeddings are determined by 2 FF-NNs with a margin objective on predicting the next word, a quasi-language modelling setting even if it utilises both the preceding and subsequent context. Multi-sense skip-gram \cite{neelakantan-etal-2014-efficient} also defines senses as cluster centroids, measuring the cosine distance of the surrounding context of $\pm 5$ words.
\newcite{li-jurafsky-2015-multi} use Chinese Restaurant Processes to decide whether to create a new cluster, and also investigate the usefulness of multi-sense embeddings in downstream tasks such as Semantic Relatedness and PoS tagging. 
\cite{chronis-erk-2020-bishop} create multi-prototype embeddings from BERT, to address word similarity and relatedness tasks. Every word \textit{w} is associated with a set of occurrence embeddings, which are computed by averaging sub-word WordPiece tokens. Senses are obtained by applying K-means clustering to the set.
Each BERT layer contributes a different set, thus the authors found that middle layers are more relevant to the similarity task, the final layers to relatedness.

\paragraph{Supervised multi-sense embeddings} Other models rely on sense-label supervision in a text corpus. \textit{context2vec} \cite{melamud-etal-2016-context2vec} builds contextual word embeddings by applying a biLSTM on text, and provides the option to create sense embeddings by using a sense-labelled corpus. 
\newcite{raganato-etal-2017-neural} frames WSD as a sequence learning problem, with the aim of finding sense labels for an input sequence. The training corpus is the same used in our work,  SemCor \cite{Miller93asemantic}, and the core architecture is a biLSTM that reads both the preceding and subsequent context. A biLSTM is also employed in LSTMEmbed \cite{iacobacci-navigli-2019-lstmembed}, 
that obtains a sense-labelled training corpus by applying the BabelFly \cite{moro-etal-2014-entity} sense tagger on the English Wikipedia and other texts. 


Recently, SenseBERT \cite{levine-etal-2020-sensebert} uses a BERT transformer encoder with two output mappings, one for the MLM-masked words, another for their WordNet supersenses.
SenseBERT relies on soft labeling, associating a masked word \textit{w} with any one of its supersenses \textit{S(w)}. Using very large text corpora is expected to reinforce the correct supersense labels.

\paragraph{KB-based methods} Sense representations can leverage WordNet glosses, as done in \newcite{chen-etal-2014-unified} after single-prototype vectors are trained with a skip-gram model.
Likewise, pre-trained word embeddings are the starting point for AutoExtend \cite{rothe-schutze-2015-autoextend}, an autoencoder architecture where word embeddings constitute the input and the target whereas the embeddings for WordNet synsets are the intermediate encoded representation.
\newcite{kumar-etal-2019-zero} use a BiLSTM and self-attention to get contextual embeddings, then disambiguate them via a dot product with sense embeddings based on WordNet definitions. 

In the last couple of years, efforts have been made to enable BERT to disambiguate between senses. GlossBERT \cite{huang-etal-2019-glossbert} takes in context-gloss pairs as input: for every target word in a context sentence, N=4 WordNet glosses are found; a classification layer determines which lemma the word assumes.
SenseEmBERT \cite{Scarlini_Pasini_Navigli_2020} relies on the BabelNet mapping between WordNet and Wikipedia pages to collect relevant text for synsets,
and computes the sense embeddings as a rank-weighted average of relevant synsets.  \newcite{scarlini-etal-2020-contexts} applies the same pipeline of of context extraction, synset embeddings and sense embeddings construction: it computes lemma representations via BERT, and uses UKB \cite{agirre-etal-2014-random} to create a set of contexts for the synset. Sense embeddings are obtained by concatenating the BERT representation of the sense contexts found in SemCor and the sense definition in WordNet.

\paragraph{Our model} We use pre-trained embeddings and WordNet glosses and relations to create a dictionary graph. The vectors of sense nodes can be viewed as sense embeddings; we are not primarily interested in their quality since our objective is not "classic" Word Sense Disambiguation or relatedness, but rather multi-sense language modelling. Future work may rely on them to improve sense disambiguation: for model variants that have to choose the correct sense among a limited number of candidates, there is an opening for the application of more complex multi-sense models than the ones explored here.

\section{Multi-Sense Language Model}

\subsection{Architecture}
\label{subsec:Architecture}

A language modelling task decomposes the probability of predicting an entire text of length $N$ as the product of each word prediction, where the probability of the next word $p(w_i)$ is influenced by the preceding context $[w_1, ..., w_{i-1}]$:
\begin{align}
    p( w_1, ..., w_N) = \prod_{i=1}^N p(w_i \vert w_1, ..., w_{i-1})
\end{align}

Our model aims to carry out two language modelling tasks: 
\begin{enumerate}[noitemsep]
    \item  Standard language modelling: 
    next-token prediction at the granularity of words.
    \item Sense prediction: next-token prediction at the granularity of WordNet senses.
\end{enumerate}

\begin{figure}[t]
    \centering
    \includegraphics[width=0.42\textwidth]{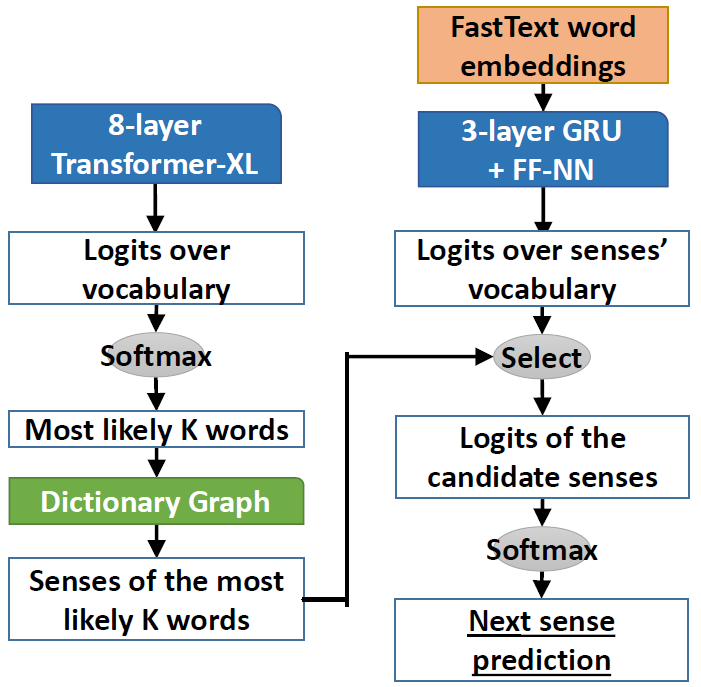}
\caption{Overview of the SelectK version of the model that uses a Transformer-XL for the StandardLM task.}
    \label{fig:model_SelectK}
\end{figure}
 
Therefore, the objective is to produce, at each position in the corpus, two probability distributions: one over the vocabulary of words and one over the vocabulary of senses.

The architecture for the standard language modelling task is either a 3-layer GRU followed by a FF-NN, or an 8-layer Transformer-XL \cite{dai-etal-2019-transformer}. Both are pre-trained on WikiText-2 \cite{WikiText:16}, and then the whole model is trained on the SemCor sense-labeled corpus \cite{Miller93asemantic}.
Some experiments use a Gold LM, that always predicts the correct next word, with the aim to examine the effectiveness of sense architectures independently from the accuracy of the standard language modelling task.

\begin{figure}[t]
    \centering
    \includegraphics[width=0.5\textwidth]{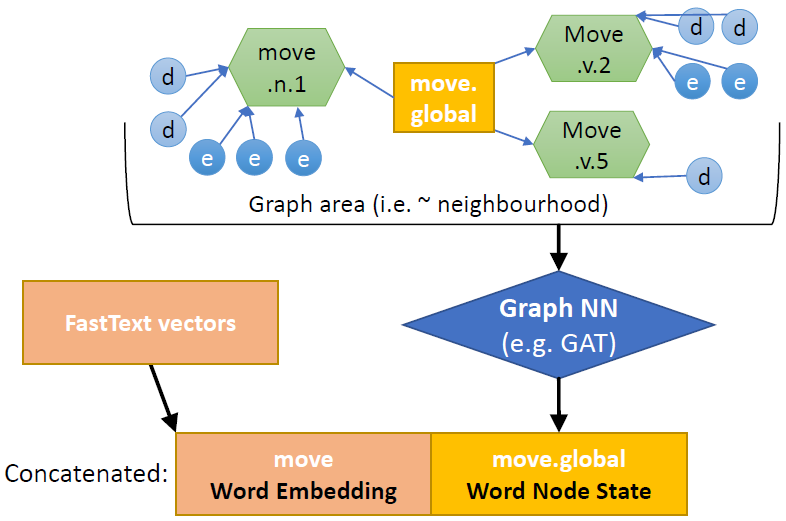}
\caption{The input signals for graph-informed sense prediction: the standard word embedding and the vector of the global node from the dictionary graph}
    \label{fig:input_signals}
\end{figure}

The input signal for the sense prediction architecture consists of FastText pre-trained embeddings \cite{bojanowski-etal-2017-enriching}, possibly augmented by concatenating the graph node corresponding to the current word, as shown in Fig. \ref{fig:input_signals}. Moreover, a model variant that uses localized prediction also relies on the K most likely next words predicted by the standard language modeling task: $ w_{t+1}^1, \ldots, w_{t+1}^K$. An example of a localized prediction variant, SelectK (§ \ref{subsec:SelectK}), is shown in Fig. \ref{fig:model_SelectK}.
~\\

The correctness of the prediction is evaluated using two measures: perplexity and accuracy.

\subsection{Dictionary Graph}

The purpose of the dictionary graph is to provide a way for the input signal to be informed by the sense embeddings
and possibly to leverage WordNet's graph structure and glosses.

First, we read in a text corpus and create the \textbf{vocabulary}. 
We register the lemmatised version of inflected forms, later connected to their parent forms.
Then, for each sense of a word, we retrieve the text of its \textbf{definitions} and \textbf{examples}, and register the connections with synonyms and antonyms.
In WordNet, senses are specified as the synsets of a word; example: $w$='bank' $\rightarrow$ [('bank.n.01'), ... ,('bank.v.07)] .

The next step is to compute the \textbf{sentence embeddings} for definitions and examples. It is possible to use any one of several methods, ranging from LSTMs to BERT's last layer. 
For the sake of computation speed, sentence embeddings are obtained as the average of the FastText word vectors.

\begin{figure}[t] 
    \centering
    \includegraphics[width=0.5\textwidth]{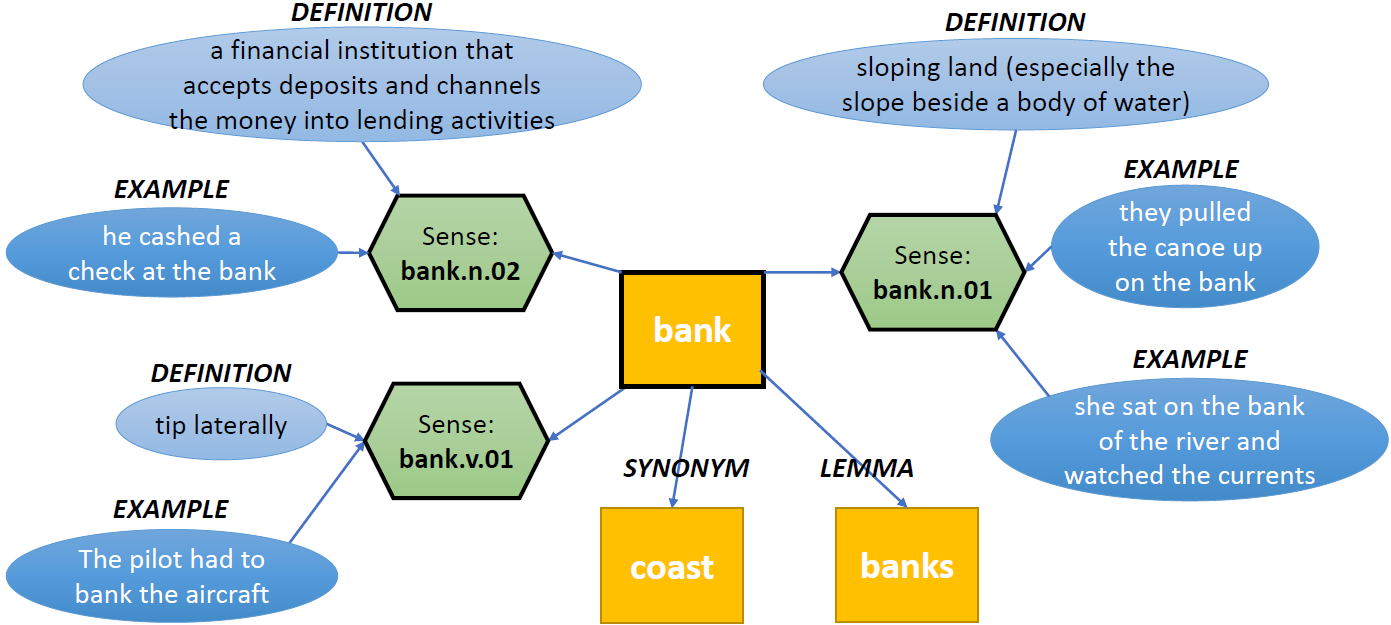}
\caption{Part of the dictionary graph for the word \textit{``bank''}. \textbf{Global} nodes are in yellow, sense nodes in green, definitions and examples in light blue and blue.}
    \label{fig:dictionary_example_graph}
\end{figure}

Finally, the nodes are initialised and stored in a graph with their edges. The graph object is created with PyTorch-Geometric \cite{pytorch_geometric:19}; before training, it holds the initial values for the sense and global nodes. The \textbf{sense node} starts as the average of the embeddings of definitions and examples for that sense. As shown in Fig. \ref{fig:dictionary_example_graph}, every sense node is directly connected to its definition and example nodes. Those determine its starting position; ideally, during training, it would be moved towards the more significant glosses. The \textbf{global node} is initialised as the FastText embedding for \textit{w}.

\subsection{Graph Attention Network}

We employ a Graph Attention Network \cite{velickovic2018graph} to update the nodes of the dictionary graph. Unlike Graph Convolutional Networks \cite{Kipf:2016tc}, a GAT does not require a fixed graph structure, and can operate on different local graph-batches like the neighbourhood of the current word. Unlike other methods like graphSAGE \cite{HamiltonYL17_graphSage} it can handle a variable number of neighbours. The underlying idea of GATs is to compute the representation vector $ h_i $ of node $ i $ based on its neighbouring nodes $ j \in \mathit{N}(i)$, which have different attention weights (i.e. importance).\\

We here describe how a GAT obtains the new state $h_i^{t+1}$ of node $i$ in a graph with $m$ nodes.
First, a linear transformation \textbf{W} is applied over all the nodes: $ \textbf{W}h_1^t , ...,  \textbf{W}h_m^t$. 
Then, for each node $j$ in the neighbourhood of $i$, we compute the non-normalised attention coefficient $e_{ij}$, via a 1-layer FF-NN with LeakyReLU activation: \begin{align}
        e_{ij}=LeakyReLU(\textbf{A}^T [\textbf{W}h_i, \textbf{W}h_j ])
\end{align}
The normalised attention coefficients $\alpha_{ij}$ are obtained by applying a softmax over the neighbourhood of i, $N(i)$. 
Finally, the new state of node $i$ is given by applying a non-linear function $\rho$ to the weighted sum of the neighbours' states: 
\begin{align}
        h_i^{t+1} = \rho \left( \sum_{j \in N(i)} \alpha_{ij}\textbf{W}h_j^{t} \right) 
\end{align}
\newcite{velickovic2018graph} report that using multiple attention heads, by averaging or by concatenating, is beneficial to stabilise the model; we therefore use 2 concatenated heads.

\subsection{Dealing with Missing Sense Labels}

Even in a sense-labelled corpus, some of the words, such as stopwords, will not have a sense label. This may cause a problem because in some versions of the model (§ \ref{subsec:SelectK}, §\ref{subsec:Sense Context Similarity}, §\ref{subsec:Self-attention coefficients}), the GRU used for the sense prediction task should be able to read the input text as an uninterrupted sequence of sense tokens.
In particular, two types of words may have no sense specification: \begin{enumerate}[noitemsep]
    \item \textbf{stopwords}: \textit{'for'}, \textit{'and'}, \textit{'of'}, etc.
 
    \item \textbf{inflected forms}: \textit{'is'}, \textit{'said'}, \textit{'sports'}, etc.
\end{enumerate}

In order to provide stopwords with a sense label, we add a \textbf{dummySense} (e.g. `for.dummySense.01') for all the words without a sense. The corresponding graph node is initialised as the single-prototype FastText vector for that word. 
Inflected forms are \textbf{lemmatised} using NLTK's WordNetLemmatizer, to be able to read and predict the senses of their parent form ('is'$\rightarrow$'be', 'said'$\rightarrow$'say' etc.)

\setlength{\tabcolsep}{2pt}
\begin{table}[t]
\centering
\fontsize{10}{10}\selectfont
\begin{tabular}{@{}ccccc@{}}
\toprule
 & \multicolumn{2}{c}{Words} & \multicolumn{2}{c}{Senses} \\
 & PPL & Accuracy & PPL & Accuracy \\ \midrule
2 GRUs & 160.94 & 0.209 & \textbf{562.46} & 0.053 \\
2 Transformer-XL & 128.99 & 0.241 & \textbf{186.72} & 0.217 \\ \bottomrule
\end{tabular}
\caption{Baselines: two separate GRUs, and two separate Transformer-XL models. Results of word and sense prediction on SemCor's test set.}
\label{Table1_separateTools}
\end{table}

\section{Architectures for Sense Prediction}

As previously described, multi-sense language modelling consists of the two tasks of standard language modelling and sense prediction, and we aim to output two probability distributions - for the next word token and the next sense token, respectively.

\subsection{GRU}
\label{subsec: GRU for senses}

A 3-layer GRU followed by a 1-layer FF-NN. There are no shared layers with the standard language modelling architecture, whether that be a Transformer-XL or another GRU. In the latter case, the 2 GRUs share the input signal, i.e. the FastText embeddings, possibly concatenated with the word node embedding. 

\setlength{\tabcolsep}{4pt}
\begin{table*}[h]
\centering
\fontsize{9.5}{9.5}\selectfont
\begin{tabular}{@{}llll|lll|lll@{}}
\toprule
\multicolumn{4}{c|}{Methods \& Parameters} & \multicolumn{3}{c|}{SemCor test} & \multicolumn{3}{c}{SensEval dataset} \\ \midrule
\begin{tabular}[c]{@{}l@{}}Senses \\ architecture\end{tabular} & \begin{tabular}[c]{@{}l@{}}Standard\\ LM\end{tabular} & K & context & \begin{tabular}[c]{@{}l@{}}Senses\\ ACC\end{tabular} & \begin{tabular}[c]{@{}l@{}}Polysem\\ ACC\end{tabular} & \begin{tabular}[c]{@{}l@{}}Globals\\ ACC\end{tabular} & \begin{tabular}[c]{@{}l@{}}Senses\\ ACC\end{tabular} & \begin{tabular}[c]{@{}l@{}}Polysem\\ ACC\end{tabular} & \begin{tabular}[c]{@{}l@{}}Globals\\ ACC\end{tabular} \\ \midrule
\textbf{MFS} & \textbf{Gold} & \textbf{-} & \textbf{-} & 0.83 & 0.62 & 1 & \textbf{0.82} & \textbf{0.62} & \textbf{1} \\
\textbf{SelectK} & \textbf{Gold} & \textbf{1} & \textbf{-} & \textbf{0.90} & \textbf{0.80} & \textbf{1} & 0.82 & 0.61 & 1 \\
SenseContext & Gold & 1 & average & 0.73 & 0.45 & 1 & 0.71 & 0.41 & 1 \\
Self-attention & Gold & 1 & average & 0.52 & 0.24 & 1 & 0.51 & 0.25 & 1 \\ \midrule
\textbf{MFS} & \textbf{TXL} & \textbf{-} & \textbf{-} & \textbf{0.24} & \textbf{0.02} & \textbf{0.24} & \textbf{0.23} & \textbf{0.04} & \textbf{0.23} \\
TXL & TXL & - &  & 0.22 & 0.02 & 0.24 & 0.20 & 0.03 & 0.22 \\
\textbf{SelectK} & \textbf{TXL} & \textbf{1} & \textbf{-} & \textbf{0.24} & \textbf{0.02} & \textbf{0.24} & 0.23 & 0.03 & 0.23 \\
SelectK & TXL & 5 & - & 0.13 & 0.01 & 0.24 & 0.11 & 0.01 & 0.22 \\
SenseContext & TXL & 1 & average & 0.22 & 0.01 & 0.24 & 0.22 & 0.02 & 0.23 \\
SenseContext & TXL & 1 & GRU & 0.19 & 0.00 & 0.24 & 0.17 & 0.00 & 0.23 \\
SenseContext & TXL & 5 & average & 0.13 & 0.01 & 0.24 & 0.12 & 0.01 & 0.22 \\
Self-attention & TXL & 1 & average & 0.17 & 0.01 & 0.24 & 0.16 & 0.02 & 0.23 \\
Self-attention & TXL & 5 & average & 0.08 & 0.01 & 0.24 & 0.07 & 0.01 & 0.23 \\ \midrule
\textbf{MFS} & \textbf{GRU} & - & - & \textbf{0.22} & \textbf{0.02}  & \textbf{0.22} & \textbf{0.22} & \textbf{0.03} & \textbf{0.22} \\
\textbf{SelectK} & \textbf{GRU} & \textbf{1} & - & \textbf{0.22} & \textbf{0.02} & \textbf{0.22} & 0.22 & 0.02 & 0.21 \\
SenseContext & GRU & 1 & average & 0.20 & 0.00 & 0.22 & 0.20 & 0.00 & 0.22 \\
Self-attention & GRU & 1 & average & 0.16 & 0.01 & 0.22 & 0.16 & 0.02 & 0.22 \\
SelectK & GRU & 5 & - & 0.13 & 0.00 & 0.21 & 0.12 & 0.01 & 0.21 \\
GRU & GRU & - & - & 0.05 & 0.00 & 0.21 & 0.06 & 0.00 & 0.21 \\ \bottomrule
\end{tabular}
\caption{The most relevant results of each method. Task: sense prediction. Datasets: SemCor's test split (10\%) and the aggregated SensEval dataset by \newcite{raganato-etal-2017-neural} .}
\label{Table_compareBest}
\end{table*}

\subsection{Transformer-XL}
\label{subsec: TXL for senses}
Transformer-XL is a left-to-right transformer encoder that operates at word-level granularity. It was chosen instead of a BERT model because the latter uses masked language modelling with WordPiece sub-word tokenisation, and thus cannot easily be adapted to a setting where the the next word and its sense should be predicted.
In our setting, the Transformer-XL 
learns word embeddings from scratch
and does not rely on the FastText vectors.

\subsection{Most Frequent Sense}
\label{subsec:Most Frequent Sense}
This heuristic baseline chooses the\textit{ most frequent sense} found in the training set for the most likely word predicted by the standard language model.

\subsection{SelectK}
\label{subsec:SelectK}
SelectK is
a localized prediction approach: first the next word is predicted, giving a set of K most likely candidates; then, the next sense is chosen among them.

As the text is read, for every location $t$, the standard language model outputs a probability distribution over the vocabulary, where the most likely K words are $ w_1, ..., w_K $. 
Every word $w_i$ has a set of senses: $ S(w_i) = \{ s_{i1}, ..., s_{iN} \}$. 
The next sense at location $t$ is chosen among the senses of the K most likely words:
\begin{align}
    s(t) \in \bigcup_{i=1}^K S(w_i)
\end{align}
A softmax function is applied over the logits of the selected senses, while all other senses are assigned a probability $\epsilon=10^{-8}$. 
The senses' logits are computed by a dedicated GRU, as described in Figure \ref{fig:model_SelectK}.
Alternatively, a Transformer-XL could be used, but it would presumably require more training data than the SemCor corpus alone.


K is a hyperparameter; K=1 means that the model always chooses among the senses of the most likely word. In general, the sense prediction performance depends on the performance of the standard language model: if all the K most likely globals are incorrect, the correct sense cannot be retrieved. We verify what happens for K=\{1,5,10\}. 

\subsection{Sense Context Similarity}
\label{subsec:Sense Context Similarity}
Another localized prediction method is to select the senses of the most likely K globals as candidates; then, rank them based on the cosine similarity between the local context and each sense's average context. Since language modelling is performed from left to right, the context is based only on the \textbf{preceding} tokens $ [w_{t-1}, ... , w_{t-c}] $ without considering subsequent tokens.

For each occurrence $s_1, ... s_N$ of a sense $s$, the occurrence context $OC(s_i)$ is computed as the average of the word embeddings of the preceding \textit{c} tokens. 
Afterwards, the Sense Context $SC(s)$ is computed as the average of the occurrences' contexts:
\begin{gather} \label{eqn:SenseContext}
    OC(s_i) = avg(w_{t-1},...,w_{t-c}) \\ \nonumber
    SC(s) = avg(OC(s_1), ... ,OC(s_N)) 
\end{gather}

We also experiment with obtaining a representation of the local context with a 3-layer GRU.

\subsection{Self-Attention Coefficients}
\label{subsec:Self-attention coefficients}
Another way to choose among the candidate senses of the most likely K globals is to use the softmax scores from the self-attention mechanism. 
Every sense $s$ has an average context $SC(s)$ it appears in, as seen in Eq. \ref{eqn:SenseContext}. 
The contexts of the candidate senses are collected in the matrix $C$. Then, a probability distribution over the senses is obtained by computing the self-attention coefficients:
\begin{gather}
    softmax\left( \frac{Q \cdot C}{\sqrt{d_k}} \right)
\end{gather}
All the rows of the query matrix $Q$ are representations of the current context. As previously, the local context can be constructed as a simple average of the last \textit{c} word embeddings, or as the output of a 3-layer GRU. 
The sense contexts in $C$ take up the role of keys in the formula of self-attention scores.

\setlength{\tabcolsep}{4pt}
\begin{table*}[]
\centering
\fontsize{9.5}{9.5}\selectfont
\begin{tabular}{@{}llll|lllr|lllr@{}}
\toprule
\multicolumn{4}{c|}{Methods \& Parameters} & \multicolumn{4}{c|}{SemCor test set} & \multicolumn{4}{c}{SensEval dataset} \\ \midrule
\begin{tabular}[c]{@{}c@{}}Senses\\ architecture\end{tabular} & \begin{tabular}[c]{@{}c@{}}Standard\\ LM\end{tabular} & K & context & \begin{tabular}[c]{@{}c@{}}Senses\\ ACC\end{tabular} & DG $\Delta$ & PPL & \multicolumn{1}{l|}{DG $\Delta$} & \begin{tabular}[c]{@{}l@{}}Senses\\ ACC\end{tabular} & DG $\Delta$ & PPL & \multicolumn{1}{l}{DG $\Delta$} \\ \midrule
SelectK & TXL & 1 & - & \textbf{0.24} & \textbf{0} & 119.6 & -0.1 & \textbf{0.23} & \textbf{0} & 172.54 & -1.34 \\
SelectK & TXL & 5 & - & \text{0.13} & \text{0} & 121.0 & 0 & \text{0.11} & \text{0} & 162.2 & +0.3 \\
SenseContext & TXL & 1 & average & 0.22 & 0 & 118.9 & +1.2 & 0.22 & 0 & 172.4 & +8.0 \\
SenseContext & TXL & 5 & average & 0.13 & 0 & \text{135.8} & \text{+0.5} & 0.12 & 0 & \text{213.2} & \text{+2.1} \\
Self-attention & TXL & 1 & average & 0.17 & +0.01 & \textbf{111.9 }& \textbf{-7.3} & 0.17 & +0.01 & \textbf{155.7} & \textbf{-16.8} \\
Self-attention & TXL & 5 & average & 0.08 & 0 & 120.6 & +1.3 & 0.08 & +0.01 & 161.7 & -4.6 \\ \midrule
GRU & GRU & - & - & \text{0.05} & \text{0} & 148.5 & -12.5 & \text{0.06} & \text{0} & 157.7 & -5.8 \\
SelectK & GRU & 1 & - & \textbf{0.22} & \textbf{0} & 131.7 & -1.6 & \textbf{0.22} & \textbf{+0.01} & \textbf{153.1} & \textbf{-3.6} \\
SelectK & GRU & 5 & - & 0.13 & -0.01 & 148.5 & -12.1 & 0.11 & -0.01 & 157.7 & -5.6 \\
SenseContext & GRU & 1 & average & 0.21 & +0.01 & 129.5 & -3.7 & 0.21 & +0.01 & 153.3 & -3.4 \\
SenseContext & GRU & 5 & average & 0.09 & +0.01 & 141.9 & -18.5 & 0.10 & +0.02 & \text{182.0} & \text{+18.7}
  \\
Self-attention & GRU & 1 & average & 0.16 & +0.01 & \textbf{128.1} & \textbf{-10.3} & 0.17 & +0.01 & 154.0 & -2.7 \\
Self-attention & GRU & 5 & average & 0.08 &	0.00 & 	139.8 & -20.7 & 0.07 & 	0.00 & 	154.8 & -9.5 \\ \bottomrule
\end{tabular}
\caption{Word and sense prediction. \textbf{DG $\Delta$} = change caused by including the input from the dictionary graph.}
\label{Table_compareWithGraphInput}
\end{table*}

\section{Evaluation}

\subsection{Dataset and Graph Settings}
\label{subsec:intro_to_eval}
To train a multi-sense language model, we need a sense-labelled text corpus. We use \textbf{SemCor} \cite{Miller93asemantic}, a subset of the Brown Corpus labelled with senses from WordNet 3.0.
Training, validation and test sets are obtained with a 80/10/10 split.
Since standard LM architectures are pre-trained on WikiText-2, the vocabulary is obtained from WikiText-2 and SemCor's training split (the latter with min. frequency=2). The dictionary graph has $\approx$169K nodes and $\approx$254K edges. Consequently, due to memory 
constraints, we apply mini-batching on the graph: for each input instance the Graph Attention Network operates on a local \textbf{graph area}. The graph area is defined expanding outwards from the global node of the current word \textit{w}.
In our experiments, a graph area contains a maximum of 32 nodes and extends for only 1 hop, thus coinciding with the \textbf{neighbourhood} of the current word's global node.

\subsection{Model variants}

In Tab. \ref{Table_compareBest} and \ref{Table_compareWithGraphInput}, we show the results of sense predictions with different architectures and parameters.\\

\textbf{Sense prediction architectures:}
\begin{itemize}[noitemsep]
    \item \textbf{GRU}: a 3-layer GRU followed by FF-NN
    \item \textbf{TXL}: an 8-layer Transformer-XL 
    \item \textbf{MFS}: Given the most likely word predicted by the standard LM, choose its most frequent sense in the training set (see § \ref{subsec:Most Frequent Sense}).
    \item \textbf{SelectK}: A GRU computes the logits over the senses' vocabulary, but the softmax is restricted to the \textit{candidate senses} only: the senses of the most likely K words from the standard LM task (see § \ref{subsec:SelectK}).
    \item \textbf{SenseContext}: Choosing among the candidate senses based on the cosine similarity of local context and average sense context (see § \ref{subsec:Sense Context Similarity}).
    \item \textbf{Self-Attention}: Choosing among the candidate senses by computing the self-attention coefficients of local context and average sense contexts (see § \ref{subsec:Self-attention coefficients}).
\end{itemize}

Tab. \ref{Table1_separateTools} compares the simplest methods available: using either two GRUs or two Transformer-XL for the two tasks of word and sense prediction. While the Transformer-XL performs better than the GRU, these are not the best results: further experiments shed light on the sense prediction model variants. 

\subsection{Results}

\paragraph{Overall Results}
In Tab. \ref{Table_compareBest}, we compare the results of the different methods for sense prediction. We report the \textbf{Senses ACC}, accuracy on the senses of all words, and the \textbf{Polysem ACC}: accuracy on the senses of polysemous words; words that have more than 1 sense are an important part of any multi-sense task, and are expected to be more difficult. 
We include the accuracy for standard word prediction, \textbf{Globals ACC}, because in our structural prediction setting, if the next word is incorrect, none of its candidate senses will be correct. 

We evaluate sense prediction using accuracy instead of perplexity, since in localized prediction methods, perplexity values are non-significant ($ \approx 10^8$), due to having assigned $\epsilon=10^{-8}$ as the probability of the non-candidate senses.

Experiments with the Gold standard language model allow us to examine the methods' capability to discriminate between senses if the next word is always predicted correctly.

The only strong accuracy values ($>$50\%) are obtained by methods using the Gold standard language model, due to the dependency on word prediction. On the SemCor test split, the best-performing method is SelectK with K=1, which is extremely reliant on the correctness of the word prediction task.
On the aggregated dataset of the SemEval-SensEval tasks \cite{raganato-etal-2017-neural}, picking the most frequent sense is a difficult baseline to beat (with Senses ACC=0.82 and Polysem ACC=0.62), but SelectK with K=1 is extremely close. 

We found that increasing K to 5 or 10 leads to a worse performance, due to the increased difficulty of choosing among a greater number of candidate senses. Using a Transformer-XL as Standard LM gives a small improvement over the GRU.
The context representation made from the average of the last 20 tokens is better than the one created by a dedicated GRU trained on SemCor.

\paragraph{Predictions for Polysemous Words}
All non-Gold model variants have very low accuracy when it comes to predicting the right sense of a polysemous word. This is due to polysemous words being the most difficult ones to predict correctly in the standard language modelling task, and therefore in a localized prediction framework.

This is corroborated by observing the word predictions of the Transformer-XL standard language model. The first batch of the SensEval dataset has 130 correct predictions out of 512 samples, where the correct words and their frequency are: 
{\fontfamily{ttfamily}\selectfont 
\{`of':10, `the':27, `$<$unk$>$':26, `and':3, `,':16, `.':10, `$<$eos$>$':16, `is':5, `to':6, 'have':1, 'a':2, `be':2, `"':1, `one':1, `'s':1, `ago':1, `with':1, `are':1\}}

Even if the Perplexity values are reasonable (with a Transformer-XL, $\sim$ 170 on SensEval and $\sim$ 120 on the SemCor test set), most polysemous words are not going to be predicted correctly.
Consequently, a way to improve multi-sense language modelling would be to utilise a better-performing model for the standard language modelling task.  
Sense prediction itself could likely be improved by training on a larger sense-annotated training dataset than SemCor. 

\paragraph{Inclusion of Dictionary Graph Input}
The input signal can be the concatenation of the FastText vector and graph node for the word \textit{w}, as shown in Fig. \ref{fig:input_signals}. This is used in two occurrences: first, by the GRU-based standard LM; secondly, by the sense architectures in the SelectK, SenseContext and Self-Attention variants. We aim to investigate whether the dictionary graph input improves the GRU-based word prediction and the local sense prediction.

Tab. \ref{Table_compareWithGraphInput} shows that the impact of the graph input signal on sense prediction is negligible, while giving a slight boost to the Self-Attention and SenseContext methods. Moreover, it often produces a small perplexity improvement for the GRU standard language model.

Future work may research how to make the graph input more helpful for word and sense prediction, by examining the use of: different Graph Neural Networks, different parameters, or different ways of encoding dictionary definitions and examples 

\section{Discussion}

As shown in Tab. \ref{Table1_separateTools}, next-token prediction at the granularity of senses is a more difficult task than standard language modelling, due to operating on a larger vocabulary with a more extensive low-frequency long tail. To try to overcome this obstacle, we proposed a localized prediction framework, finding that it is extremely reliant on the correctness of standard language model prediction. 

An argument can be made for developing better sense discrimination models that can work with a higher number of candidate senses, for instance those deriving from K=5 instead of K=1. With this in mind, we observe that there are relatively few sense-labelled datasets. The datasets organised in UFSAC format \cite{UFSAC:18} altogether contain 44.6M words, of which only 2.0M are annotated; in SemCor, 29.4\% of the tokens have a sense label. These datasets are available for English only, thus, studying the benefit of using dictionary resources for low-resource languages cannot currently be pursued until such corpora are created.

As seen in Tab. \ref{Table_compareBest}, the best results we managed to achieve are obtained by choosing among the senses of the most likely word. If a sense prediction method managed to reliably choose among a higher number of candidate senses, it would make the sense prediction task less dependant on achieving a good performance for the standard language modelling task. The question of what such a method could be remains open. It may be solved by investigating different WSD methods, and possibly by different ways of encoding the dictionary graph. Moreover, one could expect that next-token prediction, both at the word and the sense level, would benefit from a more accurate standard language modelling architecture, possibly pre-trained on a larger corpus than WikiText-2. However, WikiText-2 was chosen to avoid overwhelming SemCor's vocabulary, so this brings us back to the necessity of a larger sense-labelled corpus, that would also allow one to use an architecture different than a GRU to obtain logits over the senses.

Including the input signal from the dictionary graph only results in marginal improvements on the sense prediction task. It should be noted that the quality of the input signal is limited by the quality of the sentence encodings for the WordNet glosses, used to initialise the graph nodes. Sentence representations different from averaging FastText embeddings may achieve better results. Moreover, tuning the graph signal is surely possible, while outside of the scope of this first study of multi-sense language modelling: one could experiment with changing the size of the graph area, the number of hops and the variant of Graph Neural Network used.

\section{Conclusions}

This work constitutes the first study of multi-sense language modelling. We experiment with a localized prediction approach that predicts a word followed by a word sense; as well as learning sense representations from both its sentential context, and a sense dictionary, encoding the latter using a Graph Neural Network.

The experimental results highlight the difficulty of such a novel task. Some could regard word senses as not fit to be discretized in a language modeling task, and best represented by flexible contextual embeddings like those of transformers. We believe that specifying discrete senses in language modeling could still be improved investigating three directions: 

\begin{enumerate}[noitemsep,label=\arabic*)]
\item Training a model on larger sense-labelled resources;

\item Using different tools to build the models; for instance, creating word embeddings from BERT by averaging the WordPiece-encoded tokens, or applying other Word-sense Disambiguation methods;

\item Predicting WordNet supersenses: higher level categories such as food, artifact, person; this would avoid relying on the fine granularity of WordNet senses, solving a relatively simpler task that would still provide useful distinctions.

\end{enumerate}

Future work on this task may view it as a test bed for researching Word Sense Disambiguation, as a way of improving the precision of linking a language model to a knowledge base, or for applications such as assistive writing.


\section*{Acknowledgments}
This work is partly funded by Innovation Fund Denmark under grant agreement number 8053-00187B.

\bibliography{anthology,custom}

\begin{thebibliography}{33}
\expandafter\ifx\csname natexlab\endcsname\relax\def\natexlab#1{#1}\fi

\bibitem[{Agirre et~al.(2014)Agirre, de~Lacalle, and
  Soroa}]{agirre-etal-2014-random}
Eneko Agirre, Oier~L{\'o}pez de~Lacalle, and Aitor Soroa. 2014.
\newblock \href {https://doi.org/10.1162/COLI_a_00164} {Random walks for
  knowledge-based word sense disambiguation}.
\newblock \emph{Computational Linguistics}, 40(1):57--84.

\bibitem[{Bojanowski et~al.(2017)Bojanowski, Grave, Joulin, and
  Mikolov}]{bojanowski-etal-2017-enriching}
Piotr Bojanowski, Edouard Grave, Armand Joulin, and Tomas Mikolov. 2017.
\newblock \href {https://doi.org/10.1162/tacl_a_00051} {Enriching word vectors
  with subword information}.
\newblock \emph{Transactions of the Association for Computational Linguistics},
  5:135--146.

\bibitem[{Chen et~al.(2012)Chen, Huang, Hsieh, Kao, and
  Chang}]{chen-etal-2012-flow}
Mei-Hua Chen, Shih-Ting Huang, Hung-Ting Hsieh, Ting-Hui Kao, and Jason~S.
  Chang. 2012.
\newblock \href {https://aclanthology.org/P12-3027} {{FLOW}: A
  first-language-oriented writing assistant system}.
\newblock In \emph{Proceedings of the {ACL} 2012 System Demonstrations}, pages
  157--162, Jeju Island, Korea. Association for Computational Linguistics.

\bibitem[{Chen et~al.(2014)Chen, Liu, and Sun}]{chen-etal-2014-unified}
Xinxiong Chen, Zhiyuan Liu, and Maosong Sun. 2014.
\newblock \href {https://doi.org/10.3115/v1/D14-1110} {A unified model for word
  sense representation and disambiguation}.
\newblock In \emph{Proceedings of the 2014 Conference on Empirical Methods in
  Natural Language Processing ({EMNLP})}, pages 1025--1035, Doha, Qatar.
  Association for Computational Linguistics.

\bibitem[{Chronis and Erk(2020)}]{chronis-erk-2020-bishop}
Gabriella Chronis and Katrin Erk. 2020.
\newblock \href {https://doi.org/10.18653/v1/2020.conll-1.17} {When is a bishop
  not like a rook? when it{'}s like a rabbi! multi-prototype {BERT} embeddings
  for estimating semantic relationships}.
\newblock In \emph{Proceedings of the 24th Conference on Computational Natural
  Language Learning}, pages 227--244, Online. Association for Computational
  Linguistics.

\bibitem[{Dai et~al.(2019)Dai, Yang, Yang, Carbonell, Le, and
  Salakhutdinov}]{dai-etal-2019-transformer}
Zihang Dai, Zhilin Yang, Yiming Yang, Jaime Carbonell, Quoc Le, and Ruslan
  Salakhutdinov. 2019.
\newblock \href {https://doi.org/10.18653/v1/P19-1285} {Transformer-{XL}:
  Attentive language models beyond a fixed-length context}.
\newblock In \emph{Proceedings of the 57th Annual Meeting of the Association
  for Computational Linguistics}, pages 2978--2988, Florence, Italy.
  Association for Computational Linguistics.

\bibitem[{Devlin et~al.(2019)Devlin, Chang, Lee, and
  Toutanova}]{devlin-etal-2019-bert}
Jacob Devlin, Ming-Wei Chang, Kenton Lee, and Kristina Toutanova. 2019.
\newblock \href {https://doi.org/10.18653/v1/N19-1423} {{BERT}: Pre-training of
  deep bidirectional transformers for language understanding}.
\newblock In \emph{Proceedings of the 2019 Conference of the North {A}merican
  Chapter of the Association for Computational Linguistics: Human Language
  Technologies, Volume 1 (Long and Short Papers)}, pages 4171--4186,
  Minneapolis, Minnesota. Association for Computational Linguistics.

\bibitem[{Fey and Lenssen(2019)}]{pytorch_geometric:19}
Matthias Fey and Jan~E. Lenssen. 2019.
\newblock Fast graph representation learning with {PyTorch Geometric}.
\newblock In \emph{ICLR Workshop on Representation Learning on Graphs and
  Manifolds}.

\bibitem[{Hamilton et~al.(2017)Hamilton, Ying, and
  Leskovec}]{HamiltonYL17_graphSage}
Will Hamilton, Zhitao Ying, and Jure Leskovec. 2017.
\newblock Inductive representation learning on large graphs.
\newblock In \emph{Advances in Neural Information Processing Systems},
  volume~30, pages 1024--1034. Curran Associates, Inc.

\bibitem[{Howard and Ruder(2018)}]{howard-ruder-2018-universal}
Jeremy Howard and Sebastian Ruder. 2018.
\newblock \href {https://doi.org/10.18653/v1/P18-1031} {Universal language
  model fine-tuning for text classification}.
\newblock In \emph{Proceedings of the 56th Annual Meeting of the Association
  for Computational Linguistics (Volume 1: Long Papers)}, pages 328--339,
  Melbourne, Australia. Association for Computational Linguistics.

\bibitem[{Huang et~al.(2012)Huang, Socher, Manning, and
  Ng}]{huang-etal-2012-improving}
Eric Huang, Richard Socher, Christopher Manning, and Andrew Ng. 2012.
\newblock \href {https://aclanthology.org/P12-1092} {Improving word
  representations via global context and multiple word prototypes}.
\newblock In \emph{Proceedings of the 50th Annual Meeting of the Association
  for Computational Linguistics (Volume 1: Long Papers)}, pages 873--882, Jeju
  Island, Korea. Association for Computational Linguistics.

\bibitem[{Huang et~al.(2019)Huang, Sun, Qiu, and
  Huang}]{huang-etal-2019-glossbert}
Luyao Huang, Chi Sun, Xipeng Qiu, and Xuanjing Huang. 2019.
\newblock \href {https://doi.org/10.18653/v1/D19-1355} {{G}loss{BERT}: {BERT}
  for word sense disambiguation with gloss knowledge}.
\newblock In \emph{Proceedings of the 2019 Conference on Empirical Methods in
  Natural Language Processing and the 9th International Joint Conference on
  Natural Language Processing (EMNLP-IJCNLP)}, pages 3509--3514, Hong Kong,
  China. Association for Computational Linguistics.

\bibitem[{Iacobacci and Navigli(2019)}]{iacobacci-navigli-2019-lstmembed}
Ignacio Iacobacci and Roberto Navigli. 2019.
\newblock \href {https://doi.org/10.18653/v1/P19-1165} {{LSTME}mbed: Learning
  word and sense representations from a large semantically annotated corpus
  with long short-term memories}.
\newblock In \emph{Proceedings of the 57th Annual Meeting of the Association
  for Computational Linguistics}, pages 1685--1695, Florence, Italy.
  Association for Computational Linguistics.

\bibitem[{Kipf and Welling(2017)}]{Kipf:2016tc}
Thomas~N. Kipf and Max Welling. 2017.
\newblock {Semi-Supervised Classification with Graph Convolutional Networks}.
\newblock In \emph{Proceedings of the 5th International Conference on Learning
  Representations}, ICLR '17.

\bibitem[{Kumar et~al.(2019)Kumar, Jat, Saxena, and
  Talukdar}]{kumar-etal-2019-zero}
Sawan Kumar, Sharmistha Jat, Karan Saxena, and Partha Talukdar. 2019.
\newblock \href {https://doi.org/10.18653/v1/P19-1568} {Zero-shot word sense
  disambiguation using sense definition embeddings}.
\newblock In \emph{Proceedings of the 57th Annual Meeting of the Association
  for Computational Linguistics}, pages 5670--5681, Florence, Italy.
  Association for Computational Linguistics.

\bibitem[{Levine et~al.(2020)Levine, Lenz, Dagan, Ram, Padnos, Sharir,
  Shalev-Shwartz, Shashua, and Shoham}]{levine-etal-2020-sensebert}
Yoav Levine, Barak Lenz, Or~Dagan, Ori Ram, Dan Padnos, Or~Sharir, Shai
  Shalev-Shwartz, Amnon Shashua, and Yoav Shoham. 2020.
\newblock \href {https://doi.org/10.18653/v1/2020.acl-main.423} {{S}ense{BERT}:
  Driving some sense into {BERT}}.
\newblock In \emph{Proceedings of the 58th Annual Meeting of the Association
  for Computational Linguistics}, pages 4656--4667, Online. Association for
  Computational Linguistics.

\bibitem[{Li and Jurafsky(2015)}]{li-jurafsky-2015-multi}
Jiwei Li and Dan Jurafsky. 2015.
\newblock \href {https://doi.org/10.18653/v1/D15-1200} {Do multi-sense
  embeddings improve natural language understanding?}
\newblock In \emph{Proceedings of the 2015 Conference on Empirical Methods in
  Natural Language Processing}, pages 1722--1732, Lisbon, Portugal. Association
  for Computational Linguistics.

\bibitem[{Logan et~al.(2019)Logan, Liu, Peters, Gardner, and
  Singh}]{logan-etal-2019-baracks}
Robert Logan, Nelson~F. Liu, Matthew~E. Peters, Matt Gardner, and Sameer Singh.
  2019.
\newblock \href {https://doi.org/10.18653/v1/P19-1598} {{B}arack{'}s wife
  hillary: Using knowledge graphs for fact-aware language modeling}.
\newblock In \emph{Proceedings of the 57th Annual Meeting of the Association
  for Computational Linguistics}, pages 5962--5971, Florence, Italy.
  Association for Computational Linguistics.

\bibitem[{Melamud et~al.(2016)Melamud, Goldberger, and
  Dagan}]{melamud-etal-2016-context2vec}
Oren Melamud, Jacob Goldberger, and Ido Dagan. 2016.
\newblock \href {https://doi.org/10.18653/v1/K16-1006} {context2vec: Learning
  generic context embedding with bidirectional {LSTM}}.
\newblock In \emph{Proceedings of The 20th {SIGNLL} Conference on Computational
  Natural Language Learning}, pages 51--61, Berlin, Germany. Association for
  Computational Linguistics.

\bibitem[{Merity et~al.(2016)Merity, Xiong, Bradbury, and Socher}]{WikiText:16}
Stephen Merity, Caiming Xiong, James Bradbury, and Richard Socher. 2016.
\newblock {Pointer Sentinel Mixture Models}.
\newblock \emph{CoRR}, abs/1609.07843.

\bibitem[{Mikolov et~al.(2010)Mikolov, Karafiát, Burget, Cernocký, and
  Khudanpur}]{Mikolov:10}
Tomas Mikolov, Martin Karafiát, Lukás Burget, Jan Cernocký, and Sanjeev
  Khudanpur. 2010.
\newblock Recurrent neural network based language model.
\newblock In \emph{INTERSPEECH}, pages 1045--1048. ISCA.

\bibitem[{Miller(1995)}]{Miller95wordnet:a}
George~A. Miller. 1995.
\newblock Wordnet: A lexical database for english.
\newblock \emph{Communications of the ACM}, 38:39--41.

\bibitem[{Miller et~al.(1993)Miller, Leacock, Tengi, and
  Bunker}]{Miller93asemantic}
George~A. Miller, Claudia Leacock, Ee~Tengi, and Ross~T. Bunker. 1993.
\newblock A semantic concordance.
\newblock In \emph{Proceedings ARPA Human Language Technology Workshop}, pages
  303--308.

\bibitem[{Moro et~al.(2014)Moro, Raganato, and Navigli}]{moro-etal-2014-entity}
Andrea Moro, Alessandro Raganato, and Roberto Navigli. 2014.
\newblock \href {https://doi.org/10.1162/tacl_a_00179} {Entity linking meets
  word sense disambiguation: a unified approach}.
\newblock \emph{Transactions of the Association for Computational Linguistics},
  2:231--244.

\bibitem[{Neelakantan et~al.(2014)Neelakantan, Shankar, Passos, and
  McCallum}]{neelakantan-etal-2014-efficient}
Arvind Neelakantan, Jeevan Shankar, Alexandre Passos, and Andrew McCallum.
  2014.
\newblock \href {https://doi.org/10.3115/v1/D14-1113} {Efficient non-parametric
  estimation of multiple embeddings per word in vector space}.
\newblock In \emph{Proceedings of the 2014 Conference on Empirical Methods in
  Natural Language Processing ({EMNLP})}, pages 1059--1069, Doha, Qatar.
  Association for Computational Linguistics.

\bibitem[{Peters et~al.(2018)Peters, Neumann, Iyyer, Gardner, Clark, Lee, and
  Zettlemoyer}]{peters-etal-2018-deep}
Matthew~E. Peters, Mark Neumann, Mohit Iyyer, Matt Gardner, Christopher Clark,
  Kenton Lee, and Luke Zettlemoyer. 2018.
\newblock \href {https://doi.org/10.18653/v1/N18-1202} {Deep contextualized
  word representations}.
\newblock In \emph{Proceedings of the 2018 Conference of the North {A}merican
  Chapter of the Association for Computational Linguistics: Human Language
  Technologies, Volume 1 (Long Papers)}, pages 2227--2237, New Orleans,
  Louisiana. Association for Computational Linguistics.

\bibitem[{Raganato et~al.(2017)Raganato, Delli~Bovi, and
  Navigli}]{raganato-etal-2017-neural}
Alessandro Raganato, Claudio Delli~Bovi, and Roberto Navigli. 2017.
\newblock \href {https://doi.org/10.18653/v1/D17-1120} {Neural sequence
  learning models for word sense disambiguation}.
\newblock In \emph{Proceedings of the 2017 Conference on Empirical Methods in
  Natural Language Processing}, pages 1156--1167, Copenhagen, Denmark.
  Association for Computational Linguistics.

\bibitem[{Rothe and Sch{\"u}tze(2015)}]{rothe-schutze-2015-autoextend}
Sascha Rothe and Hinrich Sch{\"u}tze. 2015.
\newblock \href {https://doi.org/10.3115/v1/P15-1173} {{A}uto{E}xtend:
  Extending word embeddings to embeddings for synsets and lexemes}.
\newblock In \emph{Proceedings of the 53rd Annual Meeting of the Association
  for Computational Linguistics and the 7th International Joint Conference on
  Natural Language Processing (Volume 1: Long Papers)}, pages 1793--1803,
  Beijing, China. Association for Computational Linguistics.

\bibitem[{Scarlini et~al.(2020{\natexlab{a}})Scarlini, Pasini, and
  Navigli}]{Scarlini_Pasini_Navigli_2020}
Bianca Scarlini, Tommaso Pasini, and Roberto Navigli. 2020{\natexlab{a}}.
\newblock \href {https://doi.org/10.1609/aaai.v34i05.6402} {Sensembert:
  Context-enhanced sense embeddings for multilingual word sense
  disambiguation}.
\newblock \emph{Proceedings of the AAAI Conference on Artificial Intelligence},
  34(05):8758--8765.

\bibitem[{Scarlini et~al.(2020{\natexlab{b}})Scarlini, Pasini, and
  Navigli}]{scarlini-etal-2020-contexts}
Bianca Scarlini, Tommaso Pasini, and Roberto Navigli. 2020{\natexlab{b}}.
\newblock \href {https://doi.org/10.18653/v1/2020.emnlp-main.285} {With more
  contexts comes better performance: Contextualized sense embeddings for
  all-round word sense disambiguation}.
\newblock In \emph{Proceedings of the 2020 Conference on Empirical Methods in
  Natural Language Processing (EMNLP)}, pages 3528--3539, Online. Association
  for Computational Linguistics.

\bibitem[{Vaswani et~al.(2017)Vaswani, Shazeer, Parmar, Uszkoreit, Jones,
  Gomez, Kaiser, and Polosukhin}]{vaswani2017attention}
Ashish Vaswani, Noam Shazeer, Niki Parmar, Jakob Uszkoreit, Llion Jones,
  Aidan~N Gomez, {\L}ukasz Kaiser, and Illia Polosukhin. 2017.
\newblock {Attention is All you Need}.
\newblock \emph{Advances in Neural Information Processing Systems},
  30:5998--6008.

\bibitem[{Veličković et~al.(2018)Veličković, Cucurull, Casanova, Romero,
  Liò, and Bengio}]{velickovic2018graph}
Petar Veličković, Guillem Cucurull, Arantxa Casanova, Adriana Romero, Pietro
  Liò, and Yoshua Bengio. 2018.
\newblock Graph attention networks.
\newblock In \emph{ICLR}.

\bibitem[{Vial et~al.(2018)Vial, Lecouteux, and Schwab}]{UFSAC:18}
Lo{\"i}c Vial, Benjamin Lecouteux, and Didier Schwab. 2018.
\newblock \href {https://hal.archives-ouvertes.fr/hal-01718237} {{UFSAC:
  Unification of Sense Annotated Corpora and Tools}}.
\newblock In \emph{{Language Resources and Evaluation Conference (LREC)}},
  Miyazaki, Japan.

\end{thebibliography}
.

\end{document}